\documentclass[letterpaper, 10 pt, conference]{ieeeconf}
\IEEEoverridecommandlockouts                              

\newcommand{\qq}{{QuayPoints}\xspace}
\newcommand{\qpts}{{\emph{QPts}}\xspace}
\newcommand{\qpt}{{\emph{QPt}}\xspace}

\usepackage[skins,theorems]{tcolorbox}
\usepackage{graphicx}
\usepackage{multirow}
\usepackage{multicol}
\usepackage{array,booktabs}
\usepackage{bm}
\usepackage{url}
\usepackage{hyperref}
\usepackage[capitalise, noabbrev]{cleveref}
\usepackage{tabularx}
\usepackage{xcolor}
\usepackage{subfig}
\usepackage{xspace}
\usepackage[normalem]{ulem}
\usepackage[table,xcdraw]{xcolor}
\usepackage{algorithm}
\usepackage{algpseudocode}
\usepackage{amsfonts} 
\usepackage{titlesec}  
\titlespacing*{\section} {5pt}{1pt}{1pt}
\titlespacing*{\subsection} {5pt}{1pt}{1pt}

\makeatletter
\apptocmd{\@maketitle}
\makeatother

\title{\LARGE \bf
\qq{}: A Reasoning Framework to Bridge the Information Gap Between Global and Local Planning in Autonomous Racing 
}

\author{Yashom Dighe, Youngjin Kim, Karthik Dantu
\thanks{All authors are  with Department of Computer Science and Engineering, University at Buffalo, NY, 14260 USA {\tt\small \{yashomna,ykim35,kdantu\}@buffalo.edu}}%
\thanks{$^1$The name \qq{} (pronounced as keypoints) comes from abbreviating key-waypoints to KWaypoints which can be alternatively spelled as QuayPoints. Quays, pronounced as keys, are the regions of importance along the banks of a water body where boats merge/dock on land.}
}
\begin{document}

\maketitle

\begin{abstract}
Autonomous racing requires tight integration between perception, planning and control to minimize latency as well as timely decision making. A standard autonomy pipeline comprising of a global planner, local planner, and controller loses information as the higher-level racing context is sequentially propagated downstream into specific task-oriented context.
In particular, the global planner’s understanding of optimality is typically reduced to a sparse set of waypoints, leaving the local planner to make reactive decisions with limited context. This paper investigates whether additional global insights, specifically time-optimality information, can be meaningfully passed to the local planner to improve downstream decisions. We introduce a framework that preserves essential global knowledge and convey it to the local planner through QuayPoints - regions where deviations from the optimal raceline result in significant compromises to optimality. QuayPoints enable local planners to make more informed global decisions when deviating from the raceline, such as during strategic overtaking. To demonstrate this, we integrate QuayPoints into an existing planner and show that it consistently overtakes opponents traveling at up to 75\% of the ego vehicle’s speed across four distinct race tracks. 
\end{abstract}


\section{INTRODUCTION}
\label{sec:intro}
Motion planning pipelines typically consist of three key modules: a global planner that generates an ideal offline trajectory, a local planner responsible for online decision-making and trajectory generation, and a controller that executes the planned trajectory. This architecture is standard in racing stacks\cite{baumann_forzaeth_2024,betz_software_2019,alvarez_software_nodate,zang_winning_2022} where the global planner produces the optimal trajectory, commonly referred to as the {\em raceline}. This typically occurs offline and is planned for the whole track. It is then distilled into a few waypoints along the raceline and are passed to the online components of the stack. A local planner generates alternative trajectories for real-time maneuvers such as overtaking. The local planner has a short horizon and uses the raceline waypoints to merge back to the raceline after the alternative trajectory executes. Finally, a controller tracks these trajectories as closely as it can at the vehicle’s dynamics limits. This hierarchical setup is a consequence of the challenges in computing all of them at once and the need to adapt online to accommodate racing dynamics. 


The role of each module in this hierarchical setup can be seen as filtering and refining information before passing it downstream as subgoals. For instance, the global planner filters the state space to an efficient path and passes it to the local planner as a set of waypoints. The local planner, in turn, selects subgoals within a local horizon and forwards them to the controller. While the local planner and controller operate continuously in a loop, the information served as input to the global planner is essentially discarded after its offline computation job is complete.


But this raises a natural question: is a list of waypoints the most effective way for the global planner to communicate its knowledge in a competitive setting such as racing where saving  even a fraction of a second can make the difference between winning and losing? Can the global planner provide additional information for a better decision making online for the following real time scenarios that naturally occur. Here are some examples of questions we'd like to answer:
\begin{enumerate}
\item When is it not advisable to deviate from the raceline (e.g., for overtaking)?
\item If deviation is necessary, how can one merge back onto the raceline with minimal time penalty?
\end{enumerate}
\noindent{}\emph{In short, how can the local planner make globally informed decisions that minimize overall time loss while maintaining timeliness constraints of real-time operation?}

The idea of extracting important information to enable better performance in downstream modules is not new to robotics. Prior works attempt to share richer information with local planners via velocity envelopes\cite{langmann_online_2025}, stability tubes \cite{brown_safe_2017} and critical sections \cite{laurense_path-tracking_2017} but these are very planner/ controller specific and tightly coupled with specific modules in the motion planning pipeline and do not specifically address time optimality. 

In contrast, we propose a more general idea 
to introduce the idea of key-waypoints which we call \qq{} (pronounced key-points, \qpts{} for short$^1$) that are waypoints on the track that are more important to time optimality than others. 
Inspired by how professional drivers develop track intuition through practice, our approach systematically identifies \qq{} using trajectory optimization. \qq{}  distill the global planner’s knowledge into the waypoints themselves, making them directly usable for local planning without significant modifications to the logic. 
We demonstrate the effectiveness of this distillation through extensive experiments. Our results show that deviating from the raceline at a \qpt{} incurs significantly higher time penalties; up to 2 seconds compared to 0.25s on deviations at regions away from \qpts{}
Furthermore, we integrate \qq{} into a well-known local planner \cite{stahl_multilayer_2019} with minimal modifications and show that the system is able to successfully overtake opponents traveling at significantly higher speeds compared to the base local planner. Finally, our evaluation on several RoboRacer \cite{betz_f1tenth_2024} tracks demonstrates that \qq{} are not necessarily tied to corners.



\section{Related Work}
\label{sec:literature}
\subsection{Global Planning}
Global planning typically involves first solving for a spatial path (a sequence of waypoints) and then computing a velocity profile to obtain a full trajectory with temporal information \cite{lipp_minimum-time_2014}. Classical approaches include graph-based \cite{dijkstra_note_1959, stentz_optimal_1994}, sampling-based \cite{lavalle_rapidly-exploring_1998, karaman_incremental_2010, karaman_sampling-based_2011}, and geometric methods \cite{walambe_optimal_2016}. These methods, however, are mainly suited for low to moderate accelerations and are less effective in high-performance racing.

To overcome this, optimal control formulations tailored for racing are used. They fall into two groups: minimum curvature and minimum time. Minimum curvature methods \cite{braghin_race_2008, heilmeier_minimum_2020} reduce path curvature to maintain higher cornering speeds, yielding trajectories close to minimum-time results despite not explicitly optimizing for time. Minimum-time approaches \cite{casanova_minimum_2000, christ_time-optimal_2021, lot_curvilinear_2014} directly minimize lap time by jointly solving for path and speed, often via solving nonlinear optimizations or their convex approximations. These achieve better theoretical times but are computationally heavier and sensitive to model accuracy.

Data-driven alternatives have also emerged; e.g., \cite{jain_computing_2020} use Bayesian optimization to predict centerline deviations that minimize lap times.

\subsection{Local Planning, Trajectory tracking and Control}
 
To handle rapid environmental changes, several specialized local planning methods have been developed.

The first category comprises pure local planners that modify the reference path or generate a plan directly from sensor input. Typical approaches include graph search \cite{stahl_multilayer_2019}, sampling-based \cite{bak_stress_2022}, or hybrid methods \cite{ogretmen_hybrid_2023}. These rely on heuristics derived from local quantities—such as lateral deviation or obstacle distance—and lack a global notion of time optimality. They are usually paired with controllers \cite{dighe_kinematics-only_2023,freire_optimal_2022,sukhil_adaptive_2021} that track the trajectory in real time.

The second category consists of optimization-based planners that jointly compute the path and control inputs, typically using Model Predictive Control (MPC) \cite{liniger_optimization-based_2015, kalaria_local_2021}. MPC handles constraints and obstacle avoidance effectively in a receding horizon, but still optimizes only locally.

Finally, learning-based methods \cite{weiss_deepracing_2020, weiss_this_2022} directly generate agile trajectories from sensor input.

\subsection{Global-to-Local Information Channels}
Several prior works focus on richer global-to-local cues than raw way-points: velocity/ stability-envelope approaches propagate  adaptive speed profiles or  stability tubes that the local sampler or MPC must respect \cite{langmann_online_2025, brown_safe_2017}; dynamic-programming or grid planners feed convex spatial corridors or cost maps directly into the receding-horizon controller \cite{liniger_optimization-based_2015, ferguson_efficiently_2008}; “critical-section” methods flag path segments where tyre saturation is imminent and modulate control authority there \cite{laurense_path-tracking_2017}; and curvature-driven heuristics iteratively minimize path curvature to yield a low-cost line the local layer tracks \cite{kapania_sequential_2016}.

Compared to these methods that pass dense, planner-specific data downstream, we instead compress the global planner’s time-optimality insight into the waypoints by marking some of them as \qq{}. Additionally, our framework is orthogonal to all of the approaches above and can be plugged into a motion planning pipeline along with these methods.

\section{METHODOLOGY}
In this section, we first go over some preliminaries in \cref{sub-sec:prelims} and details of our trajectory generation formulation in \cref{sub-sec:formulation}. \Cref{sub-sec:keypoints} covers the procedure to identify \qq{}.
\label{sec:method}

\subsection{Preliminaries}
\label{sub-sec:prelims}
\subsubsection{Track Representation}

\begin{figure}
    \centering
    \includegraphics[trim=0 0 0 0, clip, width=\linewidth]{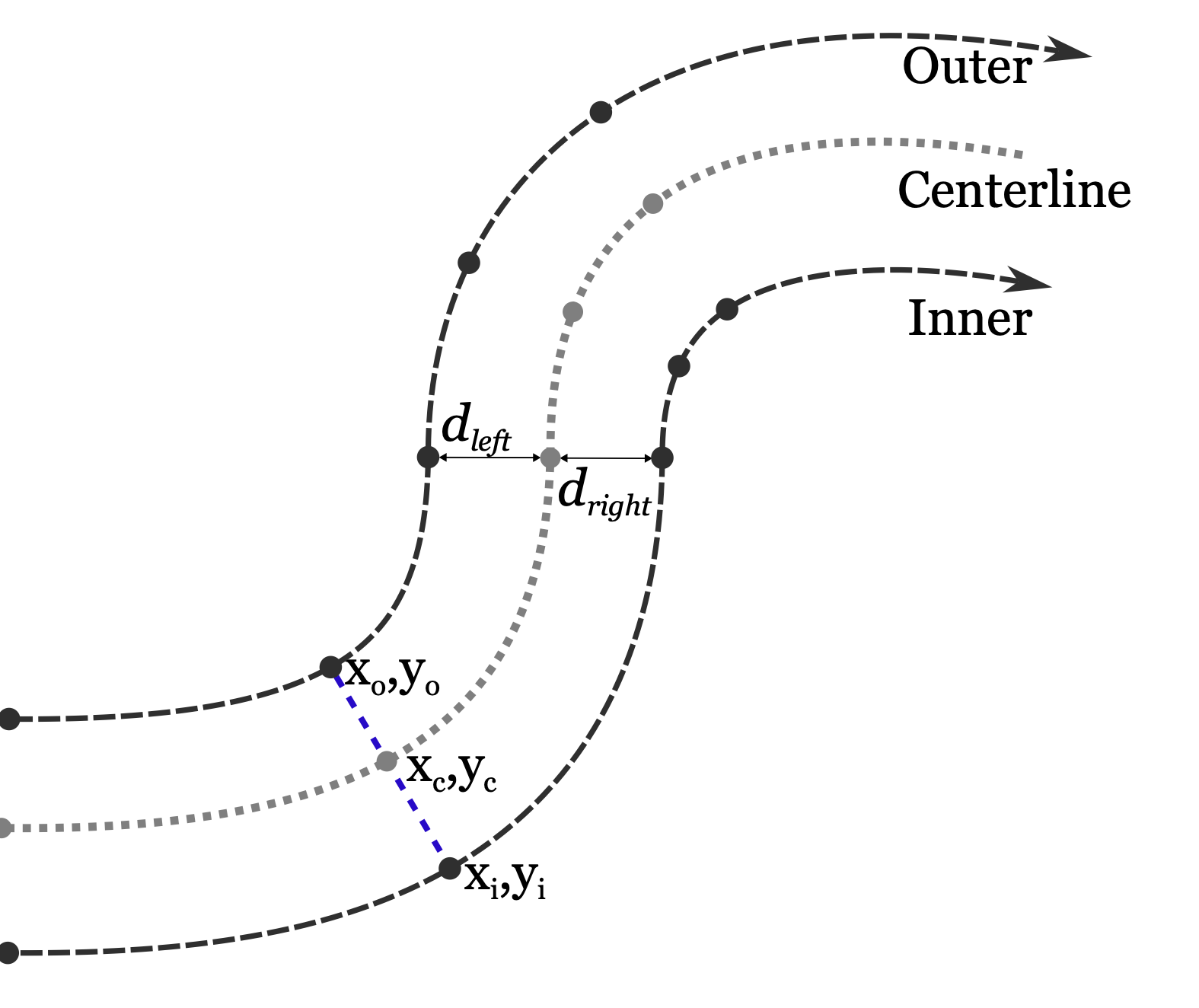}
    \caption{Track representation. The arrow specifies the direction on intended lap}
    \label{fig:track}
\end{figure}

We use the track data for scaled down tracks from \cite{betz_autonomous_2022}, which provides discretized centerline coordinates and distances to the track boundaries $(d_{left}, d_{right})$. From these, we compute inner $(i)$ and outer $(o)$ boundary points by offsetting along the centerline normal, yielding tuples $(x_o, y_o, x_c, y_c, x_i, y_i)$ of collinear points (\cref{fig:track}). Each tuple defines a line segment that we represent parametrically as
\begin{equation}
    \label{eq:convex}
    p =\lambda i + (1-\lambda)o \quad ; \quad \lambda \in [0,1] 
\end{equation} 
such that when $\lambda =0, p=o$ and when $\lambda=1, p=i$.
Using this parametric form over the traditional curvilinear coordinates in the Frenet frame, as in previous works \cite{lot_curvilinear_2014}, lets us conduct our systematic simulations (\cref{sub-sec:keypoints}) in a more intuitive manner while remaining mathematically identical. 

\subsubsection{Car Model}
We adopt the standard kinematic bicycle model \cite{rajamani_lateral_2012}, assuming rear-wheel drive, front-wheel steering, and no slip.

\begin{align} 
\dot{\textbf{q}}=\begin{bmatrix}
    \dot{x} \quad
    \dot{y} \quad
 \dot{\theta}
\end{bmatrix}^T = \begin{bmatrix}
    V\cos(\theta) \quad
    V\sin(\theta) \quad 
 \frac{V}{L}\tan(\delta) \notag
\end{bmatrix}^T 
\label{eq:bikinematic}
\end{align} \\
The input to the system is the forward linear velocity $V$, steering angle $\delta$ and $L$ is the length of the wheelbase. 

Finally, the path coordinate, $s\in[0,1]$ such that $\textbf{p(s)}\in \mathbb{R}^2$, signifies the progress along the track. 


\subsection{Trajectory Generation} 
\label{sub-sec:formulation}
To generate a raceline, we formulate a minimization problem that optimizes the total time as the objective.
\begin{align}
        T_f = \int_{0}^{t_f} dt \notag
\end{align}
However, given that $t_f$ is unknown, we use the standard change of variable trick used in prior works \cite{lipp_minimum-time_2014, verscheure_time-optimal_2009} and parameterize the objective in terms of the path coordinate $s$ such that, 

\begin{align}
        T_f = \int_{0}^{1} \frac{1}{\dot{s}}ds
\end{align}

To ensure that the trajectory runs within the boundaries of the track we use \cref{eq:convex} and treat $\lambda$ as quantity that depends on $s$.
\begin{align}\label{eq:innerOuterConvex}
    \textbf{p}_r(s) = \lambda(s)\textbf{p}_i(s) + (1-\lambda(s))\textbf{p}_o(s) \\ 
    \lambda \in [\lambda_{min}, \lambda_{max}] \quad\quad\quad\quad \notag
\end{align}
where $\textbf{p}_i(s)$ and $\textbf{p}_o(s)$ are the coordinates on inner and outer boundaries as explained previously.  
In addition to these, we have our control constraints
\begin{subequations}\label{eq:control_constraints}
    \begin{gather}
        V_{min}\leq V(t) \leq V_{max}, \quad
        \delta_{min}\leq \delta(t) \leq \delta_{max}, \\ 
         a_{min} \leq a_{long}(t) \leq a_{max}
    \end{gather}
\end{subequations} 

and finally the friction circle constraint \cite{milliken_race_1995} (using point mass dynamics) that prevents the car from slipping out of control.
\begin{align}\label{eq:friction_cric}
     a_{long}(t)^2 + a_{lat}(t)^2 \leq \mu g
\end{align}
where $\mu$ and $g$ are the coefficient of static friction and gravitational acceleration, respectively.
The quantities $V, \delta, a_{long}, a_{lat}$, can be computed using kinematic equations as 
\begin{align}
     V &= \sqrt{\dot{x} + \dot{y}}, &\delta = \text{arctan}\left(L\frac{\dot{x}\ddot{y}-\dot{y}\ddot{x}}{(\dot{x} + \dot{y})^{3/2}}\right) \notag \\
     a_{long}&=\frac{\dot{x}\ddot{x}+\dot{y}\ddot{y}}{V}, &a_{lat} =\frac{\dot{x}\ddot{y}-\dot{y}\ddot{x}}{V} \notag
\end{align}
And, $\dot{(.)}$, the time derivative, can be computed  from $(.)'$, the derivative w.r.t $s$, using chain rule. For example, $\dot{x}$ and $\ddot{x}$ can be calculated as (same for $\dot{y}$ and $\ddot{y}$)
\begin{align}
    \label{eq:dt_ds}
    \dot{x}= x'\dot{s} \quad \& \quad  \ddot{x}= x'\ddot{s} + x''\dot{s}^2 
\end{align}
Similar to \cite{lipp_minimum-time_2014}, we define 3 new variables such that 
\begin{align}
    \label{eq:def_a_b}
    b(s) := \dot{s}^2,  \quad a(s) := \ddot{s} \quad \\
    \label{eq:def_gamma}
    \gamma(s) := \lambda'' \quad\quad\quad\quad
\end{align}
From \cref{eq:def_a_b} we have,
\begin{align}
    \dot{b} &= \frac{db}{dt} = \frac{d(\dot{s}^2)}{dt} = 2\dot{s}\ddot{s} = 2\dot{s}a \notag 
\end{align}
But, by chain rule
\begin{align}
    \dot{b} &= \frac{db}{dt} = \frac{db}{ds} \frac{ds}{dt} = b'\dot{s} \notag \\
    \Rightarrow b'\dot{s} &= 2\dot{s}a \Rightarrow b' = 2a \Rightarrow b = \int 2a.ds
    \label{eq:b_a_int}
\end{align}
This transforms our objectives \cref{eq:obj_s} and time derivative calculations \cref{eq:dt_ds} to be 
\begin{align}
    \label{eq:obj_b}
    T_f &= \int_{0}^{1} \frac{1}{\sqrt{b}}ds, \\
    \dot{(x)}= x'\sqrt{b} \notag \quad &\& \quad
    \ddot{(x)}= x'a + x''b \notag
\end{align}
From \cref{eq:def_gamma}
\begin{align}
    &\lambda'' = \gamma \notag\\
    \label{eq:gamma_d}
    \Rightarrow \quad &\lambda' = \int \gamma ds \\
    \label{eq:gamma_dd}
    \Rightarrow \quad  &\lambda = \int \lambda' ds
\end{align}
where $\lambda$ is bounded as,
\begin{align}
    \lambda_{min}\leq \lambda \leq \lambda_{max} \label{eq:pathconst}
\end{align}
Putting everything together gives complete optimization problem: 
\begin{subequations}\label{eq:formuation}
    \begin{gather}
        \label{eq:formulation_obj}
        \min_{a,b, \gamma} \quad \quad  [\ref{eq:obj_b}] \\
        \label{eq:formulation_constraints}
            \textrm{s.t. :} \,\,  [\ref{eq:b_a_int}],\quad[\ref{eq:gamma_d}],\quad[\ref{eq:gamma_dd}],\quad[\ref{eq:pathconst}] ,\quad[\ref{eq:control_constraints}] \quad \& \quad[\ref{eq:friction_cric}]
    \end{gather}
\end{subequations}
We transcribe the optimization problem \ref{eq:formuation} using Hermite collocation and solve it using ipopt \cite{wachter_implementation_2006} via Casadi\cite{andersson_casadi_2019}


 

\subsection{\qq{} Extraction}
\label{sub-sec:keypoints}

Our objective is to identify locations on the track where lap time is most sensitive to the vehicle’s trajectory in order to enable our local planner to make better decisions. To achieve this, we generate a variety of racelines across the track by systematically adjusting the path constraints, then analyze their behavior to uncover consistent patterns or invariants in trajectory shapes.

Our trajectory generation method (described in \cref{sub-sec:formulation}) solves for a vector of the parameter $\lambda$ as per \cref{eq:formuation} that places a waypoint across the width of the track as per  \cref{eq:convex} for each coordinate along the track. The ideal raceline is obtained by setting the bounds on $\lambda$ to $\lambda_{min} = 0$ and $\lambda_{max} = 1$, allowing the optimizer to use the full track width. By constraining these bounds, we can generate alternate racelines that are restricted to specific sections by creating new virtual track boundaries.

\begin{figure*}
    \centering
    \includegraphics[trim=1.3cm 1.4cm 1.3cm 0cm, clip=true, width=\textwidth]{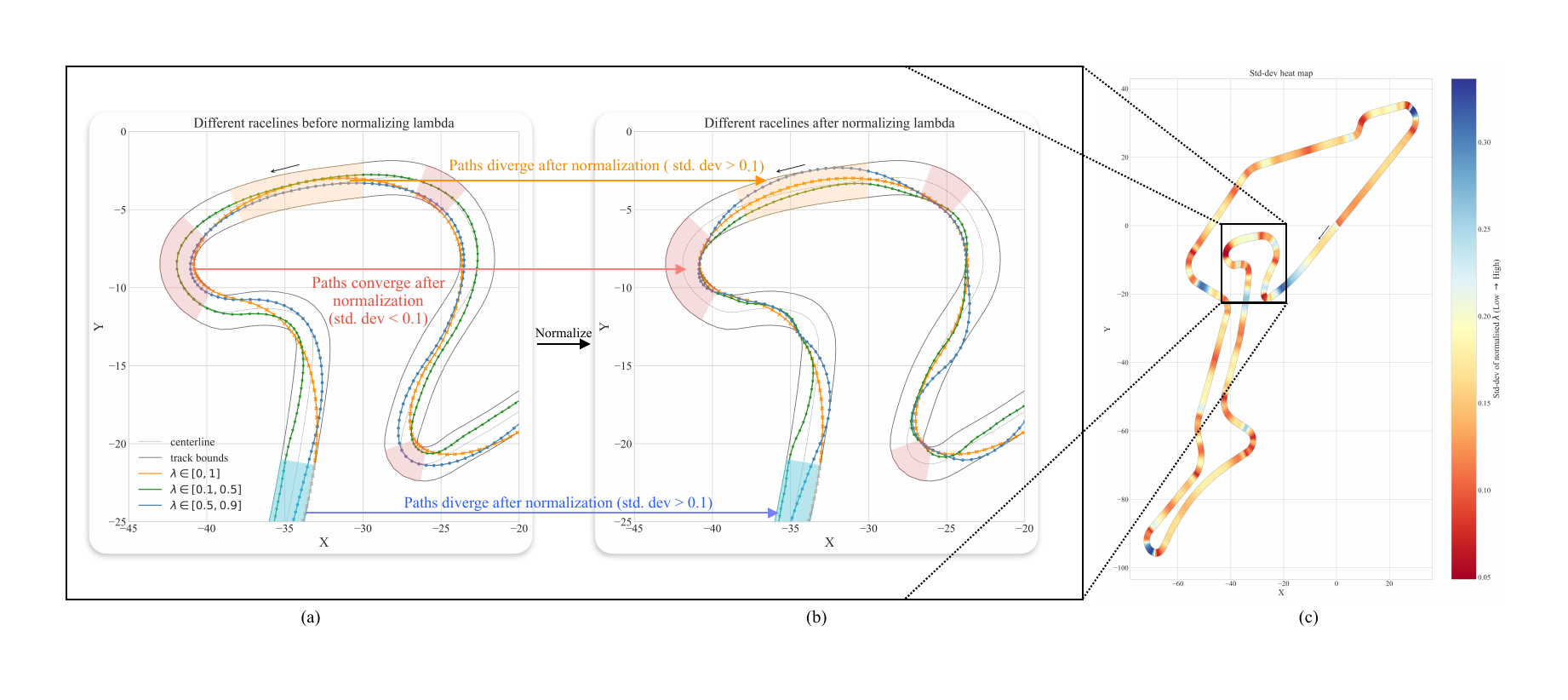}
    \caption{\textbf{\qq{} : time-critical regions revealed by invariance of lateral position across perturbed, time-optimal trajectories}. \qq{} are calculated by first solving the minimum-time trajectory multiple times while sweeping the allowable lateral position limits $(\lambda_{\min}(s),\,\lambda_{\max}(s))$ along the track \textbf{(a)} . For each solution, record the raceline’s lateral position ($lambda(s)\in[0,1]$), measured across the track from outer edge (0) to inner edge (1) at path parameter ($s$). \textbf{(b)} At each path coordinate, stack the solutions and normalize each raceline’s lateral position to a zero-to-one scale using the local lower and upper limits and measure the across-solution spread (standard deviation) of this normalized position. \textbf{(c)} Shows the std. dev $\sigma(s)$ computed from a sweep of 55 racelines for Nürburgring circuit. Thresholding ($\sigma(s)<\tau$) ($\tau=0.10$) yields QuayPoints that can guide the local planner better.}
    \label{fig:kp_heatmap}
    \vspace{-5pt}
\end{figure*}

\Cref{fig:kp_heatmap} \textbf{(a)} illustrates three such racelines: the orange line uses the full track width, while the blue and green lines are restricted to the inner and outer halves, respectively—effectively treating the centerline (black dotted line) as a virtual boundary. A key observation is that despite different constraints, all three racelines tend to pass through similar regions at certain locations. For example, they all hug the inner boundary near the first corner, switch to the outer at the next turn, and transition back to the inner as they enter the next corner (highlighted in red in).
This consistency in path shape, even under varying constraints, forms the basis of our \qq{} extraction method, outlined below

\begin{enumerate}
\item Discretize the interval $[0,1]$ into a fixed-size set of values. In this work, we discretize the interval with an 0.1 increment - ${0.0, 0.1, 0.2, \ldots, 1.0}$. 
\item Select a pair $(\lambda_{min}, \lambda_{max})$ such that $0 \leq \lambda_{min} < \lambda_{max} \leq 1$ from the discretized set.
\item Generate a trajectory using the method in \cref{sub-sec:formulation} with the selected $\lambda$ limits and store the optimal $\lambda^{*}$.
\item Repeat steps 2 and 3 for all valid $(\lambda_{min}, \lambda_{max})$ combinations from the ($N = {}^nC_{2}$) possible combinations to create a set $\mathbf{\Lambda^{*}} = \{\lambda^{*}_1, \lambda^{*}_2, \ldots, \lambda^{*}_{N}\}$ of $N$ racelines with different track limits.
\item Normalize the $\lambda^{*}$ values of each trajectory in $\mathbf{\Lambda^{*}}$ to the $[0,1]$ range using min-max normalization, producing $\mathbf{\overline{\Lambda^{*}}}= \{\overline{\lambda^{*}}_1, \overline{\lambda^{*}}_2, \ldots, \overline{\lambda^{*}}_{N}\}$.~\Cref{fig:kp_heatmap} shows the effect of this normalization (\textbf{(a)} - racelines before normalization and \textbf{(b)} - after).
\item For each track coordinate $i$, compute the standard deviation $\sigma_i$ of the normalized $\lambda$ values across all $55$ trajectories:
\begin{equation}
\sigma_i = \sqrt{ \frac{1}{N} \sum_{j=1}^{N} (\overline{\lambda^{*}_{i,j}} - \mu_i)^2 } \notag
\end{equation}
where $\overline{\lambda^{*}_{i,j}}$ is the normalized lambda value for the the way point $s = i$ on trajectory $j$ in $\mathbf{\overline{\Lambda^{*}}}$ and $\sigma_i, \mu_i$ are the corresponding mean and std. dev. of normalized lambda values for that waypoint, respectively.
\item Apply a threshold on the $\sigma_i$ values to mark a coordinate as QPt or regular waypoint.
\end{enumerate}

 Intuitively, a low $\sigma_i$ indicates that the normalized lateral positions (via $\lambda$) at that coordinate are consistent across all trajectories, regardless of the allowed $\lambda$ bounds as shown in red highlight in ~\cref{fig:kp_heatmap} \textbf{(a)} and \textbf{(b)}. This means that even when the optimizer is permitted to use a wider section of the track, it chooses not to—suggesting that the trajectory must pass through that region in a specific way to maintain time optimality. Conversely, a high $\sigma_i$ suggests that the optimizer takes advantage of the available width to adjust the path, indicating lower sensitivity to precise path placement in that region (shown in blue and orange highlight in ~\cref{fig:kp_heatmap} \textbf{(a)} and \textbf{(b)}). This implies that the shape of the normalized optimal trajectories in the low $\sigma$ regions is \emph{invariant} to the track width. Effectively, the QuayPoints anchor the trajectory and the path is shaped to go through these points regardless of track boundaries.

\noindent{} ~\Cref{fig:kp_heatmap} \textbf{(c)} visualizes the output of the method detailed above as the heatmap of the std. dev.

\section{Evaluation}
We assess \qq{} on the two questions posed in the introduction.

\emph{Do \qq{} matter for time-optimality?}
We extract \qq{}, then force small lateral deviations at matched path parameter for both \qq{} and non-\qq{} locations. Deviations at \qq{} consistently produce larger lap-time penalties, and—importantly—perturbed racelines naturally reconverge in \qq{} regions. This functions both as a principled test of time-optimality and a \emph{sanity check} on the validity of the \qq{} concept.

\emph{Do \qq{} help a local planner?}
We tag waypoints with a binary flag \((x, y, \mathrm{qp})\) and let a publicly available multilayer, graph-based local planner~\cite{stahl_multilayer_2019} preserve these time-critical regions when making short-horizon choices. This simple global-to-local signal improves performance in a practical overtaking scenario, enabling success at higher opponent speeds by avoiding costly departures at key segments.

All experiments use measured parameters from a 1{:}10 F1Tenth platform \cite{betz_f1tenth_2024}, and we tag waypoints as \qq{} when the normalized lateral-position standard deviation \(\sigma\) falls below $0.10$ (determined empirically).
\subsection{QuayPoints and Time Optimality}
\label{sub-sec:penalty}


First we verify the impact of QuayPoints on laptime by systematically blocking off regions corresponding to QuayPoints—one at a time—and solve the optimization problem described in \cref{sub-sec:formulation}. We begin by selecting a single, continuous region of QuayPoints and examining their corresponding $\lambda$ values along the raceline. We then modify the optimization constraints to restrict the feasible $\lambda$ range within that region, effectively excluding the originally optimal section of the track.

Specifically, if the $\lambda$ values for the selected QuayPoint region lie on the right half of the track ($[0.5, 1]$), we impose limits $\lambda_{i,\min} = 0$ and $\lambda_{i,\max} = 0.4$ for all points $i$ in the region. Conversely, if the values lie on the left half ($[0, 0.5]$), we use $\lambda_{i,\min} = 0.6$ and $\lambda_{i,\max} = 1$. This alteration simulates an offline obstacle that forces the solver to choose suboptimal values on the opposite side of the raceline. We repeat this process for every QuayPoint region on the track, solving the optimization each time and recording the resulting lap time. Crucially, we block only one continuous QuayPoint region at a time to simulate isolated obstacle scenarios.

As a baseline, we perform a similar procedure for non-QuayPoint regions—areas not identified as QuayPoints based on the standard deviation threshold. For each run, we randomly select a continuous region with a length sampled between the minimum and maximum lengths of QuayPoint regions. Again, only one region is blocked per run, and we record the resulting lap times. We repeat this process the same number of times as the number of QuayPoint regions on each track.

\begin{figure} [!htbp]
    \centering
    \includegraphics[trim=1.8cm 1cm 1.2cm 1.5 cm,clip,width=\columnwidth]{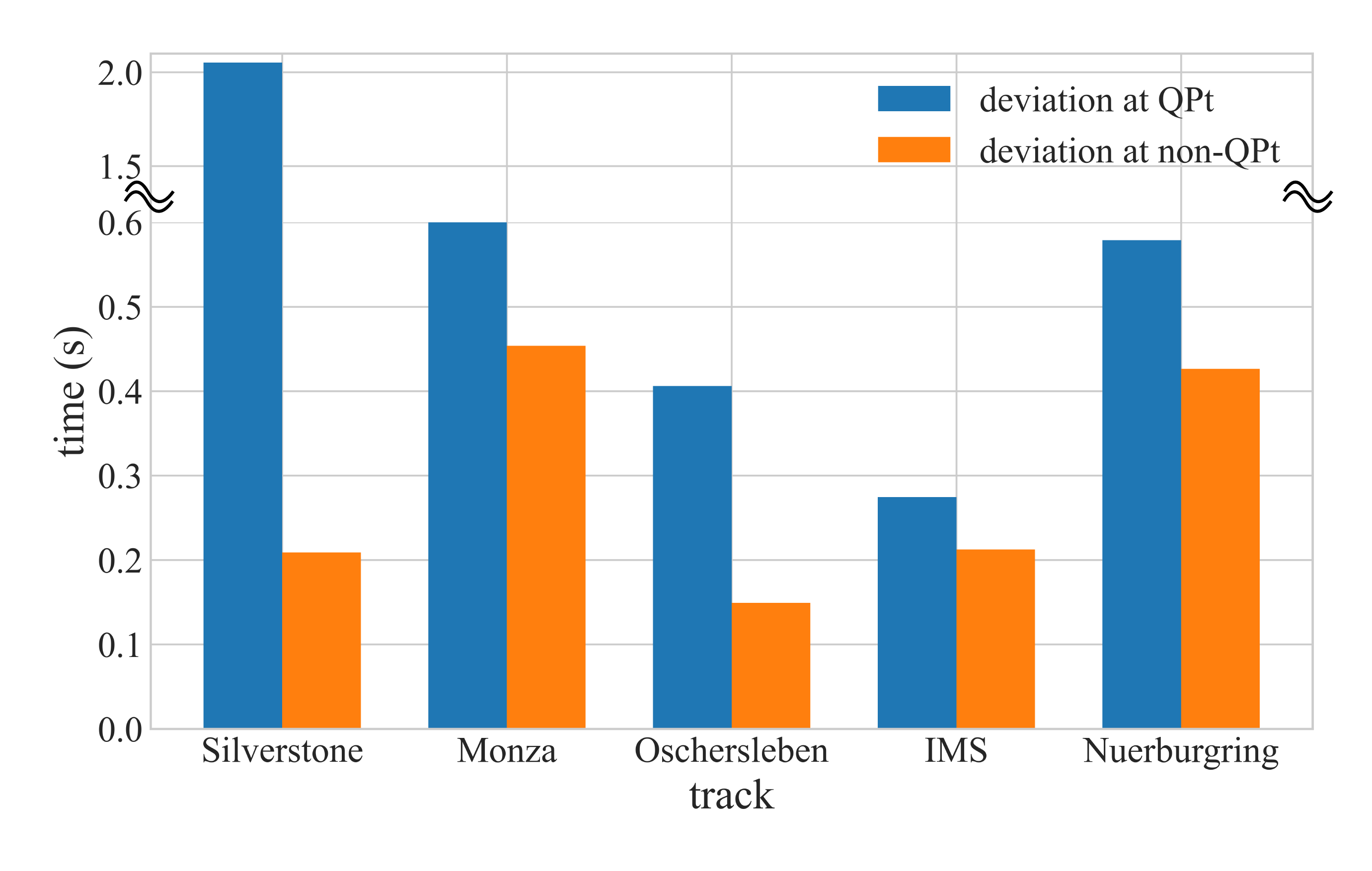}
    \caption{Average time loss if the car deviates from the raceline at a QPt vs at non-QPt. On average a single deviation at QPt results in a laptime that is 1.26\% higher to a single deviation at non-QPt which results in a 0.5\% higher laptime}
    \label{fig:time_loss}
\end{figure}

This experiment is conducted across five tracks, with results summarized in \cref{fig:time_loss}. The bar plot displays the average time penalty (i.e., the increase in lap time compared to the optimal lap time) per track when forcing the raceline to deviate at QuayPoint vs. non-QuayPoint regions. The results clearly indicate that blocking a QuayPoint region consistently incurs a higher time penalty. While the differences may appear small, they are significant in the context of motorsports like Formula 1, where finishing positions are often determined by margins as small as hundredths of a second.

\begin{figure} [!htbp]
    \centering
    \includegraphics[trim=17cm 11cm 24cm 5cm, clip, width=\columnwidth]{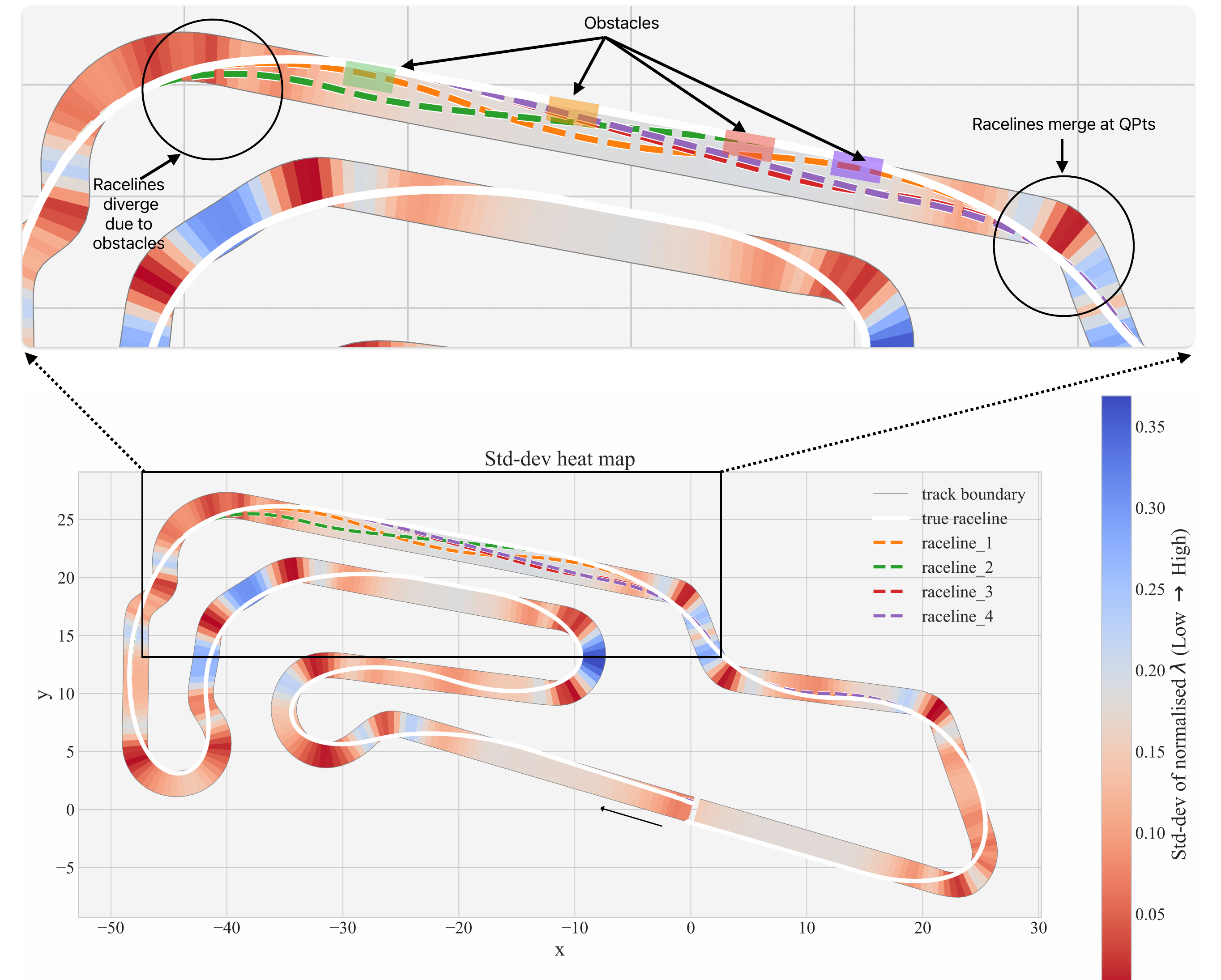}
    \caption{All deviated paths converge at the QuayPoint region (dark red), regardless of where the raceline is blocked. Obstacles of a color only exist for the raceline of the corresponding color. The white path is the raceline without any obstacles. }
    \label{fig:merge}
\end{figure}

Another trend that underscores the importance of QuayPoint regions is the observation that deviations from the raceline tend to converge and merge at these locations \cref{fig:merge}. To demonstrate this behavior, we consider a straight section of the Oschersleben racetrack and apply the same blocking strategy used previously—artificially restricting access to the optimal raceline within a localized region. This time we only block non-QuayPoint regions to analyze the resulting trajectory after an inevitable deviation.
We then run our global offline optimization multiple times, each time shifting the blocked segment progressively along the straight. The width of the blocked region is kept constant, and only one section is blocked per iteration. The resulting racelines are shown in \cref{fig:merge}. Obstacle of a given color only exist for the raceline of the same color. 

As illustrated, despite the varying positions of the blocked region, all generated paths maneuver around the obstruction and eventually converge into one at the raceline (white) at  the QuayPoint region.  Additionally, before diverging and after merging all the racelines perfectly blend into each other. This strongly suggests that QuayPoints act as anchor for the trajectory, governing its overall shape. 

\subsection{Use case: Real-time Local Planning for overtaking}
\label{sub-sec:local planning}
To demonstrate the effectiveness of our QuayPoint framework in a real racing scenario, we integrate it into a local planning method—specifically the Multilayer Graph-Based Trajectory Planning approach introduced in \cite{stahl_multilayer_2019}. Note that our contribution is not a novel overtaking method and the aim of this experiment is to demonstrate the viability of our framework and its ability to be integrated into existing methods. Correspondingly, we make minimal modifications to the planner to achieve our goal. 

We briefly summarize the baseline method here and refer the reader to the original paper for a comprehensive explanation. The core idea of the planner is to represent the track as a state lattice by placing vertices along the normal lines at fixed intervals of the raceline, indexed by the arc-length coordinate $s$. A group of vertices at a specific s-coordinate forms a layer. The spacing between layers is determined by the curvature ($\kappa$) of the raceline; layers are placed more densely in curves and more sparsely in straights. Vertices in adjacent layers are connected via splines, transforming the lattice into a graph where these splines are the edges. A pruning process removes edges based on curvature and anticipated speed loss, and the remaining edges are assigned a cost:
\begin{align}
    c_{i,j} =  s_{i,j}*(w_{len} + &w_{\kappa,av} \overline{\kappa}^2_{i,j} \\ \notag
    + &w_{\kappa,range} (\Delta \kappa_{i,j})^2 + w_{rl}|d_{lat,j}|)
\end{align}
where $s(.)$ is the spline length, $\overline{\kappa}_{i,j}$ is the average curvature of the  spline, $\Delta \kappa_{i,j}$ is the range of curvature and $d_{lat,j}$ is the lateral distance between the raceline and the vertex $j$. This process is done offline. Online, a minimum cost search is performed to generate paths and a velocity profile is generated for all. The paths are then binned into 3 actions and the planner selects which path to use.
In order to utilize QuayPoints in the local planner, we make two modifications
\begin{enumerate}
    \item \textbf{Adaptive Layer Spacing:} The distance between layers is further influenced by whether the region is a QuayPoint or not. This ensures denser sampling near QuayPoints to better capture critical trajectory variations.
    \item \textbf{Enhanced Cost Penalization at QuayPoints:} We modify the lateral deviation cost at QuayPoint layers to penalize deviations more heavily. Instead of a linear term $w_{rl}.|d_{lat,j}|$ we apply an exponential penalty $w_{rl}.e^{10|d_{lat,j}|})$ at the QuayPoints to discourage unnecessary divergence from the raceline.
\end{enumerate}
Importantly, our modifications are done only to the offline edge cost computation in the most minimal manner without touching the online graph search that actually generates the local trajectories in realtime. 
For evaluation, we adopt the same racing scenario used in \cite{stahl_multilayer_2019}, where an opponent vehicle drives along the raceline at a fraction of the ego vehicle’s maximum speed. This setup ensures that overtaking is theoretically possible. Furthermore, we utilize the same simulation that is used in the said work for their github repository. We specifically test in simulation only to evaluate the impact of the new information shared with the local planner without the influence of external disturbances such as control and localization errors which may affect the local planners decisions.

We test the approach on four tracks from the repository mentioned previously, with the opponent vehicle starting 10 meters ahead of the ego vehicle. We have tested a suite of scenarios and present the most interesting results. We vary the opponent’s maximum speed from 60\% up to the speed where both methods fail in 5\% increments. For each setting, we simulate 5 races of three laps each, across four different tracks, and record whether the ego vehicle successfully overtakes the opponent. 

\begin{figure} []
    \centering
    \includegraphics[trim=45cm 16cm 23cm 12cm, clip, width=\columnwidth]{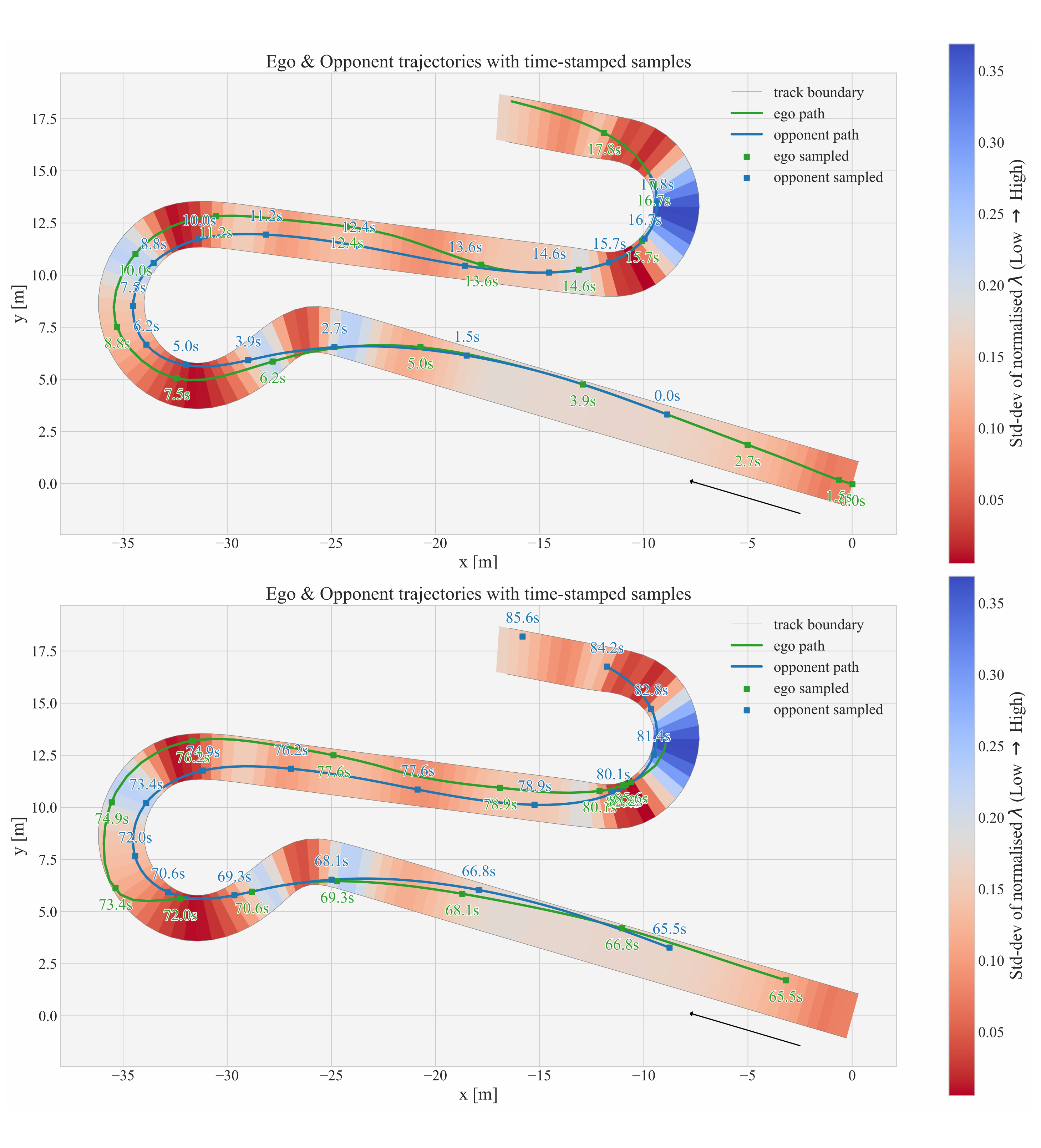}
    \caption{Overtake attempts by kp aware method (Top) and without Kp (Bottom).}
    \label{fig:overtake}
\end{figure}

\begin{figure} []
    \centering
    \includegraphics[trim=0.35cm 0 1.2cm 0cm, clip, width=\columnwidth]{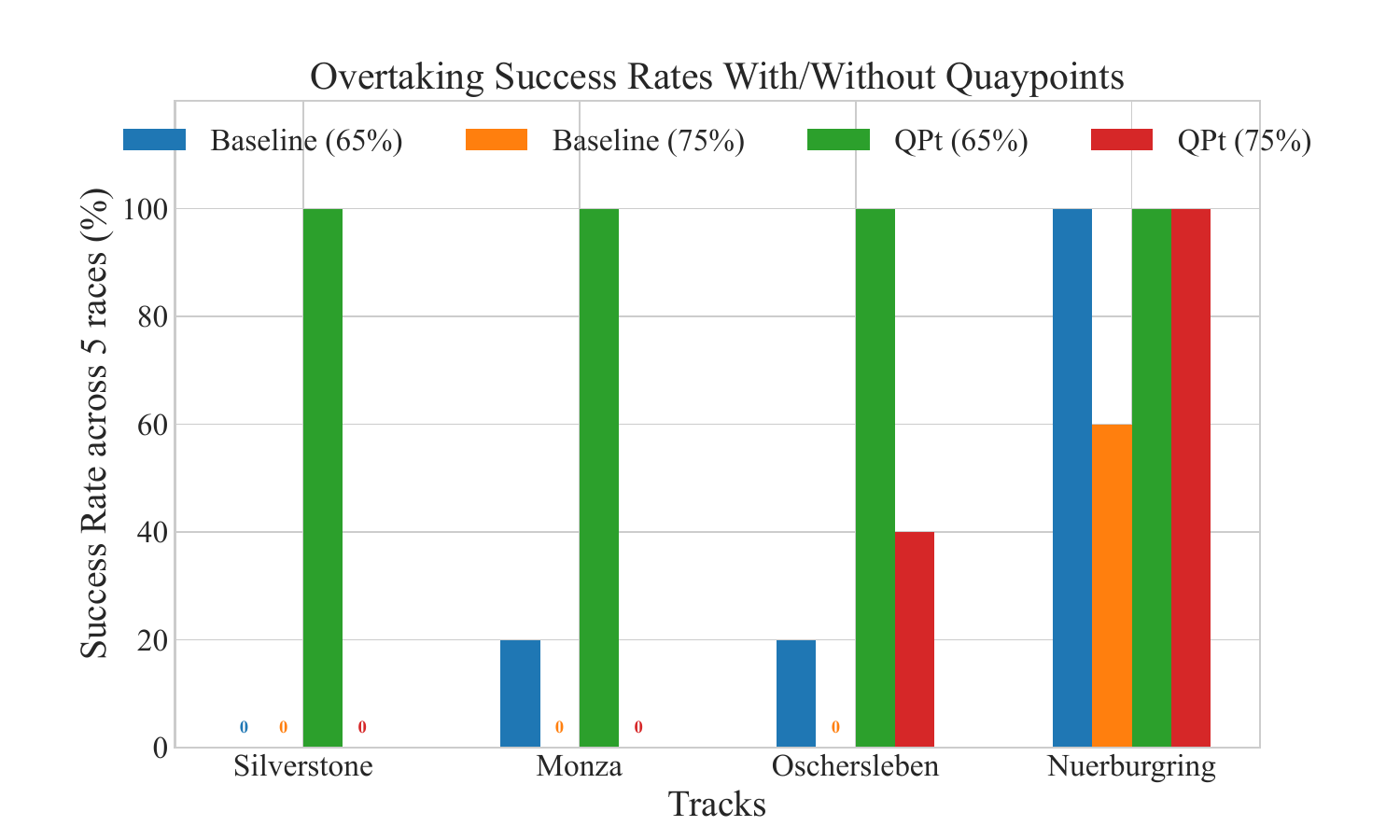}
    \caption{Success rates of overtaking maneuvers with and without Quaypoints across four tracks at two opponent speed settings (65\% and 75\% of the ego vehicle’s speed). Each group of four bars corresponds to a track, comparing the vanilla method (blue and orange bars) with Quaypoint-enhanced method (green and red bars). Results show that Quaypoints substantially improve overtaking consistency, particularly at higher opponent speeds. For opp speeds $<$ 65\% both methods reliably succeed}
    \label{fig:overtake_bars}
\end{figure}

\Cref{fig:overtake_bars} summarizes the experiment. We list only the speeds where the two methods diverge—below these both succeed reliably, and above them both fail due to acceleration limits, not path choice. Up to 65 \% of top speed, either approach can overtake an opponent. Beyond that, the baseline’s success rate drops.  This performance gain can be understood by considering the local planner’s tradeoff between spatial clearance and maneuver time. In the baseline method, large deviations from the raceline to gain space around the opponent often leads to slower trajectories (demonstrated in ~\cref{fig:overtake}-bottom). If the deviation occurs in QuayPoint regions — where the optimal path is highly sensitive — this time loss becomes significant, preventing successful overtakes.
In contrast, the QPts-aware planner avoids large deviations in QuayPoint regions (demonstrated in ~\cref{fig:overtake}-top). This leads to more efficient overtaking maneuvers without sacrificing overall lap time, demonstrating the value of QuayPoint integration.
At 75 \% we found that the QPt-aware runs followed identical paths, yet some still failed, indicating the issue may be in the velocity–profile generation rather than the path itself.
Once again, it is worth highlighting that this enhanced behavior emerges without altering any other aspect of the original planning pipeline - the trajectory sampling process, candidate evaluation, and selection mechanisms remain completely unchanged. 

\section{Discussion}

\subsection{Advantages and potential use cases}
{\bf Beyond Time Optimality}: In essence, our framework assigns importance to waypoints based on a heuristic (time optimality in our case), making the idea very general. It may, in principle, be paired with \emph{any} global trajectory generator that employs an optimality criterion — minimum jerk, minimizing lane changes, energy efficiency, or passenger comfort — and \emph{any} local planner or controller without significantly modifying their internal logic. For example, in highway driving, minimizing lane changes might be the optimality criterion to ensure safety. In urban driving, smoothness metrics such as minimum jerk or acceleration bounds could drive the selection. 

{\bf Real-world applications}: Because \qq{} are extracted \emph{post-hoc} from whatever raceline (or trajectory) the current global planner produces, they can be consumed by the local layer in many forms — cost shaping (as we demonstrate), hard constraints, candidate filtering, or heuristic bias — without imposing requirements on the planner’s structure. The concept is therefore \emph{planner-agnostic}: (i) it overlays seamlessly on existing envelopes or corridors, (ii) adds virtually no online computation, and (iii) preserves the guarantees and reactivity of the host planners while injecting principled information. This generality is particularly appealing given that autonomous vehicle companies such as Waymo and Zoox already record terabytes of trajectory and ride data daily. Mining such data for invariant regions across drivers, environments, or heuristics could allow large-scale, data-driven extraction of \qq{} that reflect not only time optimality but also comfort, safety, or traffic-efficiency metrics. Thus, \qq{} provide a plug-and-play reasoning mechanism for closing the global-to-local information gap with minimal overhead and maximal compatibility.

\begin{figure}
    \centering
    \includegraphics[width=\linewidth]{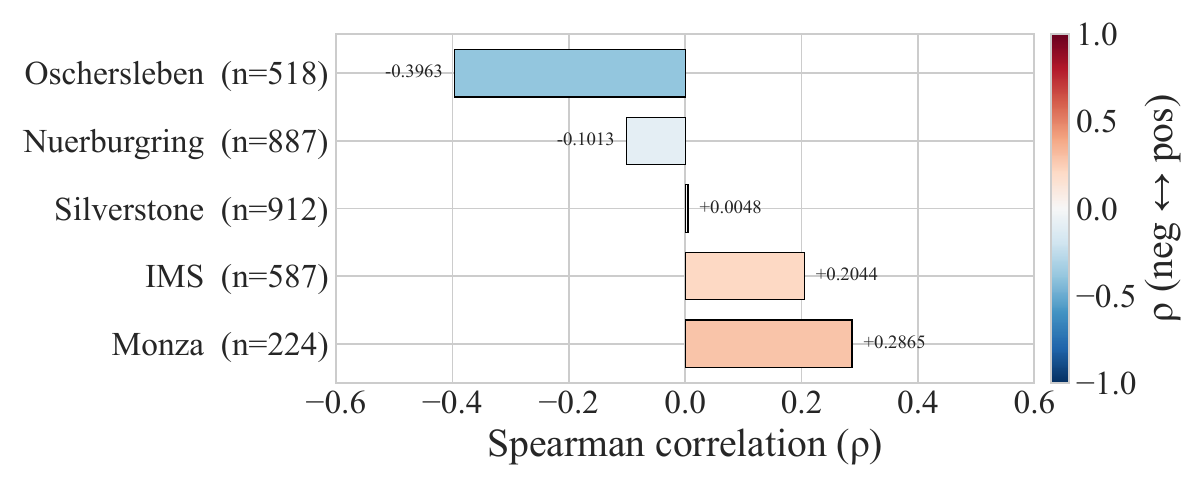}
    \caption{Correlation between curvature ($\kappa$) and the std.dev of lambda ($\lambda$) }
    \label{fig:corr}
\end{figure}

{\bf Correlation with curvature}: \cref{fig:merge} shows the Spearman correlation of \qq{} with curvature across five tracks. Looking at \cref{fig:kp_heatmap,fig:merge}, we can clearly see that  \qq{} are not confined to corners — they frequently appear on long straights, while corner regions exhibit high variance. Quantitatively, the Spearman correlation between curvature and the normalized lambda standard deviation (\cref{fig:corr}) is negative on Oschersleben, positive on Monza and IMS, and weak on Nürburgring and Silverstone. Taken together, these track-specific patterns indicate that curvature-based heuristics for applications such as corner strategies may not apply uniformly across circuits; instead, \qq{} offer a simple, data-driven basis for reasoning about the track during race preparation.

Beyond this, \qq{} can also address the common question of choosing horizon size—based on the convergence and merge shown in \cref{fig:merge}, as a form of supervision in learning-based overtaking methods or to assist in reducing the dimensionality of the global planning problem for faster convergence. They can also be used for selecting nodes in methods such as \cite{jain_computing_2020} instead of manual selection or linearly spaced points on the centerline.

\subsection{Limitations}
QuayPoint extraction is inherently iterative and involves evaluating many candidate trajectories. Depending on track length, exploring all possible combinations of limits on lambda can currently take 2–6 hours per track on a modern laptop. While our early attempts to link \qq{} to geometric quantities for a simplified extraction procedure have been inconclusive, this opens an exciting avenue for future research. Studying their occurrence in controlled path geometries—such as isolated straights, perfect circles, and simple S-curves—may reveal predictive relationships that could eliminate the need for exhaustive trajectory computations.


Secondly, our current approach treats \qq{} as binary: a region is either a QPt or it is not. Introducing a graded ranking of \qq{} importance could empower local planners to make even more informed decisions, prioritizing critical sections for maximum performance. 

\section{CONCLUSION}
\label{sec:concl}
This work shows that a small set of track \qq{} — detected as regions with consistent normalized lateral placement across perturbed, time-optimal racelines — can act as a compact, planner-agnostic channel for passing global time-optimality downstream. 
Across five tracks, forcing deviations at QuayPoints consistently incurred larger lap-time penalties than comparable deviations at non-QuayPoints, and perturbed racelines naturally reconverged in QuayPoint regions. Integrated with a multilayer graph-based local planner using only minimal, offline changes, QuayPoint awareness improved overtaking success at higher opponent speeds by avoiding costly departures in time-critical segments.
Taken together, these results motivate richer global-to-local channels—incorporating graded (non-binary) QuayPoint salience, learned one-shot predictors, and perturbations over control as well as state limits—to better generalize across vehicles and tracks. We envision that our method will serve as a platform upon which future methods can iterate and improve upon.

\bibliographystyle{ieeetr}
\bibliography{references2}

\end{document}